\documentclass[conference]{IEEEtran}
\IEEEoverridecommandlockouts
\usepackage{cite}
\usepackage{amsmath,amssymb,amsfonts}
\usepackage{algorithmic}
\usepackage{graphicx}
\usepackage{textcomp}
\usepackage{xcolor}
\usepackage{caption} 
\usepackage{hyperref}
\usepackage{caption}   % 使 caption* 有效
\usepackage{adjustbox} % 调整表格大小

\usepackage{booktabs} 
\def\BibTeX{{\rm B\kern-.05em{\sc i\kern-.025em b}\kern-.08em
    T\kern-.1667em\lower.7ex\hbox{E}\kern-.125emX}}
\begin{document}

\title{BYOCL:  Build Your Own Consistent Latent with Hierarchical Representative Latent Clustering}

\author{
    \IEEEauthorblockN{Jiayue Dai\textsuperscript{\textdagger, 1}, Yunya Wang\textsuperscript{\textdagger, 1}, Yihan Fang\textsuperscript{\textdagger,1}, Yuetong Chen\textsuperscript{\textdagger,1}, Butian Xiong\textsuperscript{1 2*}}
    
    \IEEEauthorblockA{
        \textit{Chinese University of Hong Kong, Shenzhen\textsuperscript{1} , Infused Synapse AI\textsuperscript{2}}\\
        \{jiayuedai, yunyawang, yihanfang, yuetongchen, butianxiong\}@link.cuhk.edu.cn
    }
    \thanks{\textsuperscript{\textdagger} These authors contributed equally and are considered co-first authors.}
    \thanks{\textsuperscript{*} Corresponding author.}
}

% \author{
%     \and
%     \IEEEauthorblockN{Jiayue Dai\IEEEauthorrefmark{1}}
%     \IEEEauthorblockA{
%         \textit{Chinese University of Hong Kong, Shenzhen}\\
%         jiayuedai@link.cuhk.edu.cn
%     }
%     \and
%     \IEEEauthorblockN{Yunya Wang\IEEEauthorrefmark{1}}
%     \IEEEauthorblockA{
%         \textit{Chinese University of Hong Kong, Shenzhen}\\
%         yunyawang@link.cuhk.edu.cn
%     }
%     \and
%     \IEEEauthorblockN{Yihan Fang\IEEEauthorrefmark{1}}
%     \IEEEauthorblockA{
%         \textit{Chinese University of Hong Kong, Shenzhen}\\
%         yihanfang@link.cuhk.edu.cn
%     }
%     \and
%     \IEEEauthorblockN{Yuetong Chen\IEEEauthorrefmark{1}}
%     \IEEEauthorblockA{
%         \textit{Chinese University of Hong Kong, Shenzhen}\\
%         yuetongchen@link.cuhk.edu.cn
%     }
%     \and
%     \IEEEauthorblockN{Butian Xiong\IEEEauthorrefmark{1}}
%     \IEEEauthorblockA{
%         \textit{Chinese University of Hong Kong, Shenzhen}\\
%         butianxiong@link.cuhk.edu.cn
%     }
% }

\maketitle

\begin{abstract}

%\begin{multicols}{2}
To address the semantic inconsistency issue with SAM or other single-image segmentation models handling image sequences, we introduce BYOCL. This novel model outperforms SAM in extensive experiments, showcasing its Hierarchical prototype capabilities across CLIP and other representations. BYOCL significantly reduces time and space consumption by dividing inputs into smaller batches, achieving exponential time reduction compared to previous methods. Our approach leverages the SAM image encoder for feature extraction, followed by Intra-Batch and Inter-Batch clustering algorithms. Extensive experiments demonstrate that BYOCL far exceeds the previous state-of-the-art single image segmentation model. Our work is the first to apply consistent segmentation using foundation models without requiring training, utilizing plug-and-play modules for any latent space, making our method highly efficientModels are available at \href{https://github.com/cyt1202/BYOCL.git}{https://github.com/cyt1202/BYOCL.git}.

\end{abstract}
% \begin{IEEEkeywords}
% component, formatting, style, styling, insert.
% \end{IEEEkeywords}

\section{Introduction}

\label{sec:intro}
\begin{figure}[htbp]
\centering
\includegraphics[width=\linewidth]{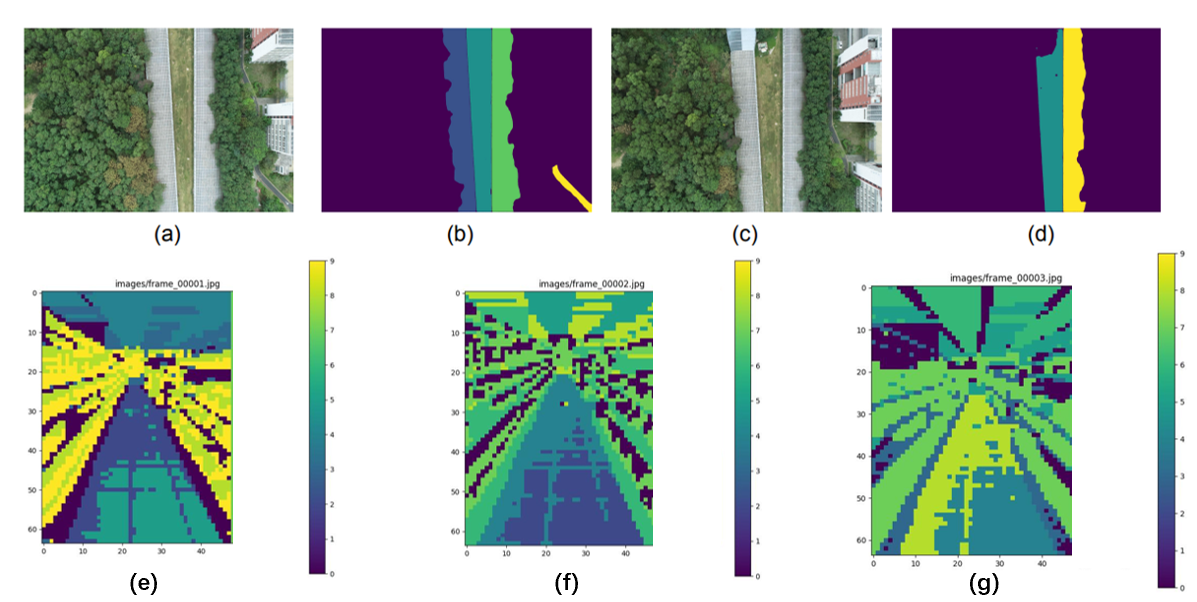}
\caption{\label{fig: inconsistancy}image(a) and image(c) are real-world scenes, which are continuously captured pictures. images (b) and image(d) are SAM-segmented results which show the inconsistency problem. Images (e) (f) (g) are inconsistent SAM-segmented results of the grocery-store dataset.}
\end{figure}

% Large Language Models (LLMs), when scaled and pre-trained on broad data with self-supervision, demonstrate substantial progress in zero-shot and few-shot generalization across various NLP tasks.\cite{bommasani2021opportunities}. 

\begin{figure}[htbp]
\centering
\includegraphics[width=0.7\linewidth]{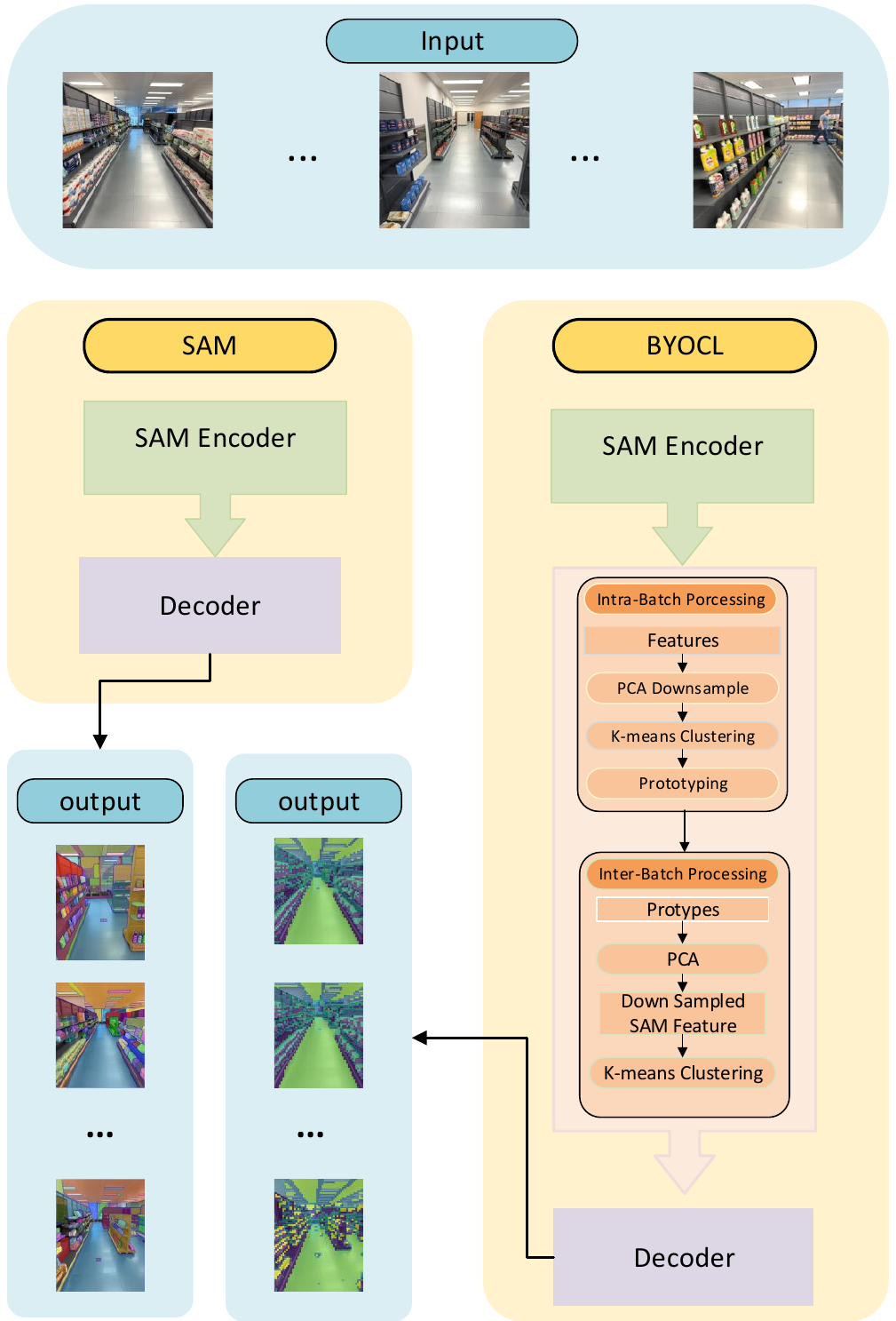}
\caption{\label{fig: introduction} Based on SAM image encoder, our method(BYOCL) adds intra-batch clustering and inter-batch clustering algorithms. After the decoder, we get segmented pictures that are semantically consistent. As shown in the graph, the results demonstrate noticeable improvements in semantic consistency compared with SAM.}
\end{figure}

\begin{figure*}[htbp]
\centering
\includegraphics[width=\linewidth]{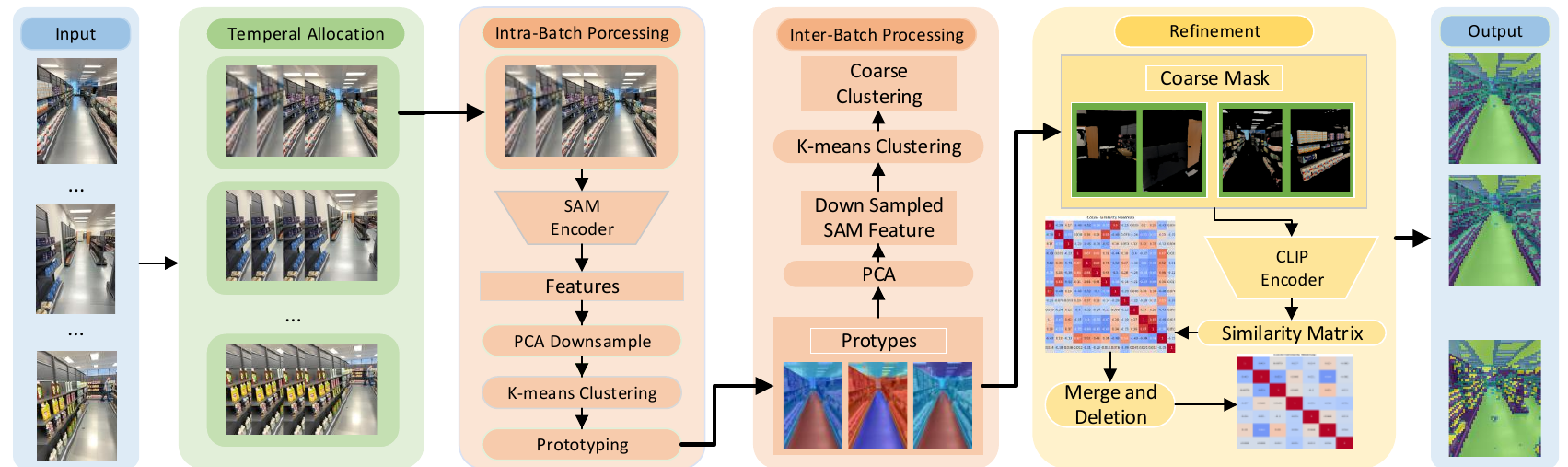}
\caption{\label{fig:top-image}This figure is a detailed description of our method. After we input a sequence of images in our model, these images are tiled in batches with the batch size = 4. Following the coarse-to-fine logic, we first design an Intra-Batch Processing which is composed of a SAM encoder, PCA Downsample, K-means Clustering and Prototyping. SAM Encoder here is used to extract image features.PCA is used to reduce the dimensionality of the features. k-means method is used to cluster the reduced feature vectors. After extract the prototype which is the cardinal feature vector of each group, we then propose Inter-Batch Processing part and input the
prototypes into the PCA and K-means clustering.The output will be shown in \ref{fig:cluster2}.}
\end{figure*}

Large Language Models (LLMs), when scaled and pre-trained on broad data with self-supervision, demonstrate strong zero-shot and few-shot generalization across NLP tasks \cite{bommasani2021opportunities}. Similarly, although the Segment Anything Model (SAM) \cite{kirillov2023segment} excels in image segmentation, it struggles with semantic inconsistency across varied images among a sequence, leading to unreliable segmentation and hindering downstream tasks (Figure \ref{fig: inconsistancy}).

To address this issue, we propose BYOCL for image and video segmentation by utilizing a Hierarchical Clustering Method. Our output is more consistent than SAM image segmentation(Figure \ref{fig: introduction}).

BYOCL involves intra-batch processing and inter-group clustering. We first perform intra-batch processing by batching neighboring pictures, extracting the features, and applying PCA Processing and K-means clustering. Then, we conduct inter-batch processing on these batches and visualize the results.
Detailed descriptions are illustrated in Section \ref{method}. Compared with other segmentation models, our method uncovers the underlying interrelation among different scenes and ensures that segmentation results remain consistent across different images of the same area.

Moreover, we have conducted extensive experiments on various segmentation method such as SAM and SAM2, employing different datasets(MOSE, DAVIS)and metrics (IOU, F1, recall)to establish a reliable benchmark. 
Our contributions to this work are summarized as follows:
\begin{itemize}
\item We propose a novel zero-shot segmentation model (BYOCL) built upon the SAM image encoder to alleviate the semantic inconsistency problem. Our model can identify the interrelation among different pictures when applied to varied datasets.
\item We perform the state-of-the-art comparison on challenging benchmarks with diverse domains.
\item The experiments on various datasets demonstrate the effectiveness of our model on open-set image segmentation tasks.
\end{itemize}
In the following sections, we will discuss related work in the SAM model and introduce the fundamentals of how the Segment Anything Model works. We begin with a detailed account of our model methodology. Subsequently, we introduce temporal allocation, intra-batch processing, inter-batch processing, refinement. Finally, we present our experiment and results on different datasets and metrics.

\section{Related work}
\label{sec:Related}
\subsection{Segment and Track Anything Models}
Deva\cite{Cheng_2023_ICCV}, SAM-Track\cite{cheng2023segment}and Track Anything Model(TAM)\cite{yang2023track}integrate SAM model with advanced Video Object Segmentation (VOS) models (such as  XMem \cite{cheng2022xmem}),  to achieve interactive tracking and segmentation in videos.
These models use SAM for mask initialization and refinement, and VOS models are used for handling mask adjustment and tracking tasks. However, these approaches face limitations, such as poor mask propagation quality due to domain gaps. Instead of building an interactive pipeline, BYOCL focuses on uncovering the interrelation among different images by utilizing the SAM image encoder for feature extraction and applying Intra-batch and inter-batch clustering.

\subsection{SAM 2: Segment Anything in Images and Videos}
Segment Anything Model 2 (SAM2) \cite{geetha2024sam} is a foundational model designed to address the challenge of promptable visual segmentation across both images and videos. Built upon a streamlined transformer architecture, SAM2 incorporates a streaming memory mechanism that facilitates real-time processing of video data. This model excels in accurately segmenting objects within individual images and efficiently managing multi-frame segmentation to track dynamic scene changes in videos. Moreover, SAM2 offers automatic image segmentation, allowing for adaptive detection and segmentation of objects without the need for manual annotations. Its advanced segmentation capabilities and seamless integration of multiple tasks make SAM2 highly versatile and applicable across various domains in computer vision. 

\subsection{Matching Anything by Segment Anything}
The matching Anything by Segment Anything (MASA)\cite{li2024matching} model achieved outstanding performance in multiple object tracking (MOT) tasks. By leveraging rich object segmentation from the SAM model, MASA learns instance-level correspondence from various data transformations\cite{masa}. Moreover, the MASA adapter, a tracking adapter, can enhance models' performance in video tracking tasks when integrated with foundational models like SAM and GroundingDINO\cite{liu2023grounding}. By utilizing segmentation and detection models, MASA improves feature tracking capability. Our work focuses on a different direction. While MASA focuses on tracking video features, BYOCL aims to solve inconsistency problems in segmentation tasks. BYOCL uncovers the interrelation among different images and segments features across diverse domains with zero-shot foundation models. 

\section{Method}\label{method}

In this section, we will mainly introduce a detailed description of the research process and the design of the experiments.The detailed flowchart of our approach is in Figure \ref{fig:top-image}.

\subsection{Temporal Allocation}

We begin by inputting a dataset of various images from a grocery store.

To extend contrastive learning from the instance level to the batch level, we tile n images into batches with the batch size equals to 4, ensuring each batch contains an equal number of images.

\subsection{Intra-Batch Processing}
\begin{figure}[htbp]
\centering
\includegraphics[width=0.60\linewidth]{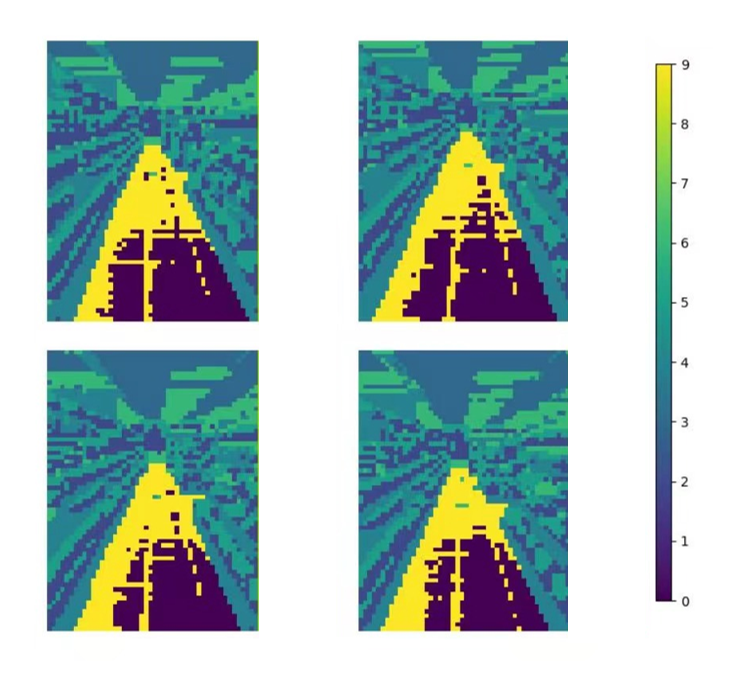}
\caption{\label{fig:firstclustering}This clustering result is an example outcome of Intra-batch clustering step.The results of segmentation within a batch are consistent.}
\end{figure}

We employ the SAM image encoder and embedding techniques to extract features from each batch of images. This process generates feature vectors of 256 dimensions for each pixels in images, constructing a feature space. The output of this feature extraction is a four-dimensional array------(batch-size, height, weight, feature vectors), that is (4, 64, 64, 256) in our method.
The features matrix of all image batches is input into a Principal Component Analysis (PCA) model to reduce the dimensionality of the features to 20. Subsequently, the k-means clustering method is employed to cluster the reduced feature vectors. 
Figure \ref{fig:firstclustering} in this section is a visualization outcome of one batch for an example.
% \href{https://www.overleaf.com/learn/how-to/Including_images_on_Overleaf}{including images on Overleaf}.

\subsection{Inter-Batch Processing}
Then we extract the prototype which is the cardinal feature vector of each group. Here the prototype is set as the mean value of the members in each group, having the dimension of (n/4*k*256) through all batches. Iteratively, we input the prototypes into the PCA and K-means clustering just like the module in the intra-batch processing.
\begin{itemize}
\item Prototype Evaluation. For the feature vectors that were not previously processed by PCA, evaluate the prototype of each cluster(the means of each cluster of features is taken as a prototype).
\par For each cluster \( j \), the prototype \( \mathbf{c}_j \) can be obtained by calculating the mean of all data points within that cluster. 
\item PCA Processing. The obtained prototype matrix is processed with dimensionality reduction.
\item Prototype K-means Clustering Processing. Clustering the prototype matrix by the k-means method after dimension reduction. Finally, every extracted feature can correspond to the category after k-means clustering in a list. Apply these mappings to each image and visualize them.
\end{itemize}

\begin{figure}[htbp]
\centering
\includegraphics[width=0.85\linewidth]{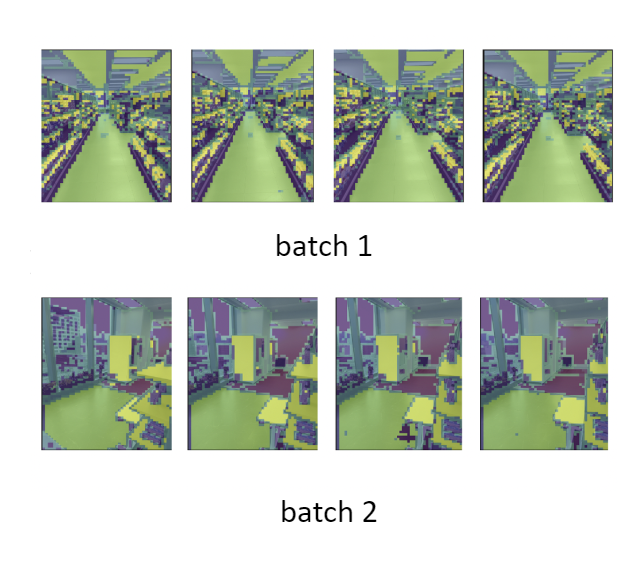}
\caption{\label{fig:cluster2}This clustering result is an example outcome of inter-batch clustering step.The segmentation results of different photos in the same scene prove the consistency of BYOCL. }
\end{figure}

%Table~\ref{tab:widgets}, 
Figure \ref{fig:cluster2} is the example outcome of the Inter-batch visualization, which improves the segment inconsistency. 

We also provide a coarse to fine stratgy to refine the rough mask result of Inter-batch processing.
%please see this help article on %\href{https://www.overleaf.com/learn/latex/tables}{tables}. 

%第二次可视化结果

\section{Result}
We have a certain amount of visualization results. The first part is What the visualization of Intra-batch clustering looks like, and The second part is how the Inter-batch clustering should look like. We also provide a coarse to fine stratgy to refine the rough mask result we have (64*64) and achieve a better segment result.

\subsection{visualization of Intra-batch clustering}
The outcome is visualization results in groups(batch-size). However, visual results in the segmentation of different groups of images are not guaranteed segmentation consistency. The sample output looks like what is shown in Figure \ref{fig:firstclustering}. 
\subsection{visualization of Inter-batch clustering} 
The outcome is consistent visual results of all image segmentation. The sample output looks like what is shown in Figure \ref{fig:cluster2}.We also refine the rough segmentation result from coarse to fine stratgy to further optimize Our results.

\begin{table*}[ht]
\centering
\caption{Comparison of methods on \textbf{DAVIS} and \textbf{MOSE} datasets.}
\label{tab:davis_mose}
\begin{adjustbox}{max width=\textwidth}
\begin{tabular}{@{}lccc|ccc|cccc|ccc@{}}
\toprule
\textbf{Dataset} & \multicolumn{6}{c}{\textbf{DAVIS}} & \textbf{} & \multicolumn{6}{c}{\textbf{MOSE}} \\ 
\cmidrule(lr){2-7} \cmidrule(lr){8-14}
\textbf{Method} & \multicolumn{3}{c}{\textbf{SAM}} & \multicolumn{3}{c}{\textbf{BYOCL}} & & \multicolumn{3}{c}{\textbf{SAM}} & \multicolumn{3}{c}{\textbf{BYOCL}} \\ 
\cmidrule(lr){2-4} \cmidrule(lr){5-7} \cmidrule(lr){9-11} \cmidrule(lr){12-14}
sequence & IOU$\uparrow$ & F1$\uparrow$ & Recall$\uparrow$ & IOU$\uparrow$ & F1$\uparrow$ & Recall$\uparrow$ & sequence & IOU$\uparrow$ & F1$\uparrow$ & Recall$\uparrow$ & IOU$\uparrow$ & F1$\uparrow$ & Recall$\uparrow$ \\ 
\midrule 
bike-packing & 0.3070 & 0.4333 & 0.3136 & 0.4165 & 0.5823 & 0.7277 & 013103f6 & 0.5562 & 0.5742 & 0 & 0.4324 & 0.4564 & 0 \\
boat & 0.6227 & 0.8328 & 0.7419 & 0.6541 & 0.8154 & 0.9934 & 02deca50 & 0.4342 & 0.5872 & 0.7419 & 0.4406 & 0.5925 & 0.8153 \\
dogs-jump & 0.5259 & 0.6847 & 0.5290 & 0.5333 & 0.5871 & 0.5433 & 08746283 & 0.8050 & 0.8799 & 0.8110 & 0.8430 & 0.9125 & 0.8920 \\
longboard & 0.5981 & 0.7508 & 0.6051 & 0.7767 & 0.8681 & 0.7856 & 0c13e1e7 & 0.7293 & 0.8360 & 0.7304 & 0.8005 & 0.8864 & 0.9474 \\
disc-jockey & 0.5562 & 0.6937 & 0.5576 & 0.5644 & 0.7204 & 0.5701 & 1106f3a7 & 0.7095 & 0.8096 & 0.7157 & 0.7599 & 0.8595 & 0.9235 \\
\midrule Avg & 0.5220 & 0.6705 & 0.5494 & \textbf{0.5890} & \textbf{0.7147} &\textbf{ 0.7240} & Avg & 0.6468 & 0.7373 & 0.5998 & \textbf{0.6552} & \textbf{0.7414} &  \textbf{0.7156 }\\
\bottomrule
\end{tabular}
\end{adjustbox}
\vspace{0.2cm}
\caption*{Table \ref{tab:davis_mose}: This table compares metrics between SAM and BYOCL methods on selected sequences from both DAVIS and MOSE datasets. Each row represents a sequence, and the columns display metrics: IOU, F1, and recall. Overall, BYOCL shows better performance on both datasets.}
\end{table*}

\section{Experiment}
\subsection{Introduction to Experiments}

\par Extensively evaluating BYOCL, we conducted experiments on three datasets: a non-open dataset, an open-source dataset based on Davis benchmark, and an open-source dataset based on Mose.We compared the segmentation accuracy between our method and SAM on each dataset.
\par The SAM (Segment Anything Model 2) \cite{geetha2024sam} approach has gained attention due to its simplicity and adaptability for both image and video segmentation tasks. The SAM uses a straightforward CNN architecture to extract spatial features from individual frames, optimizing for high spatial resolution and boundary accuracy. 
\par In the following experiments, we conduct a comprehensive evaluation of our proposed BYOCL method against SAM. Through these experiments, we aim to demonstrate the effectiveness of our method in achieving higher segmentation accuracy and consistency, as measured by metrics such as mean Intersection over Union (IoU), F1-score and recall.

% \begin{enumerate}
%     \item \textbf{Dataset.} All of our experiments are built on the MOSE open source dataset. MOSE (Multi-Object Segmentation Evaluation) dataset\cite{ding2023mose} is a benchmark dataset specifically designed for evaluating segmentation algorithms in scenarios involving multiple objects.  It was curated to address the challenges in real-world applications, where objects of varying scales, shapes, and appearances coexist in complex backgrounds. The MOSE dataset aims to provide a comprehensive evaluation platform that includes diverse and challenging segmentation tasks, making it suitable for testing the robustness and accuracy of both image and video segmentation models.
%     \item[] \textit{\textbf{Data Type}}: Contains thousands of annotated images and hundreds of video sequences.
%     \item[] \textit{\textbf{Resolution}}: Typically ranges from 720p (1280x720 pixels) to 1080p (1920x1080 pixels).
%     \item[] \textit{\textbf{Annotations}}: Provides detailed pixel-level ground truth masks for precise evaluation.
%     \item \textbf{Environment and Tools.} The server is equipped with 88 x NVIDIA GeForce RTX 3090 Gpus, each with 24GB of video memory, using NVIDIA Driver Version 555.42.06 and CUDA version 12.5. 
%     We used Python 3.9.19 as the primary programming language. The main libraries used in the experiment and their versions are NumPy 1.26.4, Pillow 10.4.0, tqdm 4.66.5, matplotlib 3.9.1, argparse and so on.
% \end{enumerate}
\subsection{Data Description}
\par The grocery store dataset, which is not yet annotated with ground truth, is used solely for visualization purposes. In contrast, for the DAVIS and MOSE datasets, we selected several specific sequences as test samples. This selection was based on the fact that other sequences contained indistinct objects or exhibited minimal contrast between objects and their backgrounds. Each sequence consists of hundreds of frames, with each frame representing a slightly varied pose of a single object within a video.

\subsection{Evaluation Metrics}
\par We employed the following evaluation metrics to assess segmentation performance: the average Intersection over Union (IoU), the average F1-score, and the average recall. These metrics collectively offer a comprehensive evaluation of both the regional accuracy and boundary precision of the segmentation models.

\subsection{Results and Analysis}
\par The visualizations of the segmentation results on the grocery store dataset are presented in Figure \ref{fig:firstclustering}. Due to camera motion, all images in each batch exhibit slight variations. Nonetheless, we are able to maintain segmentation consistency by accurately segmenting the same object across the different frames. In terms of the evaluation metrics, across both the DAVIS and MOSE datasets, our proposed method, BYOCL, achieves superior performance in terms of mean IoU, mean F1-score, and mean recall in Table \ref{tab:davis_mose}, demonstrating improved segmentation accuracy and consistency.

\par Moreover, our method is significantly more time-efficient compared to SAM. For instance, while SAM requires several hours to segment the DAVIS and MOSE datasets, BYOCL completes the segmentation process within a single hour.

\section{Conclusion}
\par Our method, BYOCL, comprises several components, including intra-batch processing, inter-batch processing, and refinement. This architecture effectively addresses the issue of segmentation inconsistency in SAM, particularly in cases involving semantically continuous images. Additionally, the refinement process not only sharpens object boundaries but also reduces computational time by limiting segmentation to a single image. Our experimental results demonstrate that BYOCL outperforms SAM in both segmentation accuracy and time efficiency when processing semantically continuous images. However, BYOCL does face challenges in multi-object segmentation, where it exhibits less capability compared to SAM.

\section*{Acknowledgment}

% The preferred spelling of the word ``acknowledgment'' in America is without 
% an ``e'' after the ``g''. Avoid the stilted expression ``one of us (R. B. 
% G.) thanks $\ldots$''. Instead, try ``R. B. G. thanks$\ldots$''. Put sponsor 
% acknowledgments in the unnumbered footnote on the first page.
Most of the equipment of the current research is founded
by the following institutes:
\begin{itemize}
\item Future Network of Intelligence Insitute, the Chinese
University of Hong KongShenzhen(FNII)
\item School of Science and Engineering, the Chinese University of Hong Kong, Shenzhen
\end{itemize}

\newpage
\bibliographystyle{IEEEtran}
\bibliography{main}

\begin{thebibliography}{10}
\providecommand{\url}[1]{#1}
\csname url@samestyle\endcsname
\providecommand{\newblock}{\relax}
\providecommand{\bibinfo}[2]{#2}
\providecommand{\BIBentrySTDinterwordspacing}{\spaceskip=0pt\relax}
\providecommand{\BIBentryALTinterwordstretchfactor}{4}
\providecommand{\BIBentryALTinterwordspacing}{\spaceskip=\fontdimen2\font plus
\BIBentryALTinterwordstretchfactor\fontdimen3\font minus \fontdimen4\font\relax}
\providecommand{\BIBforeignlanguage}[2]{{%
\expandafter\ifx\csname l@#1\endcsname\relax
\typeout{** WARNING: IEEEtran.bst: No hyphenation pattern has been}%
\typeout{** loaded for the language `#1'. Using the pattern for}%
\typeout{** the default language instead.}%
\else
\language=\csname l@#1\endcsname
\fi
#2}}
\providecommand{\BIBdecl}{\relax}
\BIBdecl

\bibitem{bommasani2021opportunities}
R.~Bommasani, D.~A. Hudson, E.~Adeli, R.~Altman, S.~Arora, S.~von Arx, M.~S. Bernstein, J.~Bohg, A.~Bosselut, E.~Brunskill \emph{et~al.}, ``On the opportunities and risks of foundation models,'' \emph{arXiv preprint arXiv:2108.07258}, 2021.

\bibitem{kirillov2023segment}
A.~Kirillov, E.~Mintun, N.~Ravi, H.~Mao, C.~Rolland, L.~Gustafson, T.~Xiao, S.~Whitehead, A.~C. Berg, W.-Y. Lo \emph{et~al.}, ``Segment anything,'' in \emph{Proceedings of the IEEE/CVF International Conference on Computer Vision}, 2023, pp. 4015--4026.

\bibitem{Cheng_2023_ICCV}
H.~K. Cheng, S.~W. Oh, B.~Price, A.~Schwing, and J.-Y. Lee, ``Tracking anything with decoupled video segmentation,'' in \emph{Proceedings of the IEEE/CVF International Conference on Computer Vision (ICCV)}, October 2023, pp. 1316--1326.

\bibitem{cheng2023segment}
Y.~Cheng, L.~Li, Y.~Xu, X.~Li, Z.~Yang, W.~Wang, and Y.~Yang, ``Segment and track anything,'' \emph{arXiv preprint arXiv:2305.06558}, 2023.

\bibitem{yang2023track}
J.~Yang, M.~Gao, Z.~Li, S.~Gao, F.~Wang, and F.~Zheng, ``Track anything: Segment anything meets videos,'' \emph{arXiv preprint arXiv:2304.11968}, 2023.

\bibitem{cheng2022xmem}
H.~K. Cheng and A.~G. Schwing, ``Xmem: Long-term video object segmentation with an atkinson-shiffrin memory model,'' in \emph{European Conference on Computer Vision}.\hskip 1em plus 0.5em minus 0.4em\relax Springer, 2022, pp. 640--658.

\bibitem{geetha2024sam}
A.~S. Geetha and M.~Hussain, ``From sam to sam 2: Exploring improvements in meta's segment anything model,'' \emph{arXiv preprint arXiv:2408.06305}, 2024.

\bibitem{li2024matching}
S.~Li, L.~Ke, M.~Danelljan, L.~Piccinelli, M.~Segu, L.~Van~Gool, and F.~Yu, ``Matching anything by segmenting anything,'' in \emph{Proceedings of the IEEE/CVF Conference on Computer Vision and Pattern Recognition}, 2024, pp. 18\,963--18\,973.

\bibitem{masa}
------, ``Matching anything by segmenting anything,'' \emph{CVPR}, 2024.

\bibitem{liu2023grounding}
S.~Liu, Z.~Zeng, T.~Ren, F.~Li, H.~Zhang, J.~Yang, C.~Li, J.~Yang, H.~Su, J.~Zhu \emph{et~al.}, ``Grounding dino: Marrying dino with grounded pre-training for open-set object detection,'' \emph{arXiv preprint arXiv:2303.05499}, 2023.

\end{thebibliography}

\end{document}